\documentclass{article}
    \PassOptionsToPackage{numbers, compress}{natbib}

\usepackage[final]{neurips_2021_preregistration}

\usepackage[utf8]{inputenc} 
\usepackage[T1]{fontenc}    
\usepackage{hyperref}       
\usepackage{url}            
\usepackage{multirow}
\usepackage{booktabs}       
\usepackage{amsfonts}       
\usepackage{nicefrac}       
\usepackage{microtype}      
\usepackage{xcolor}         
\usepackage{authblk}
\usepackage[utf8]{inputenc}
\usepackage{graphicx}
\usepackage{caption} 
\usepackage{algorithm2e}

\title{Beyond Flatland: \\
Pre-training with a Strong 3D Inductive Bias}

\author{\centering
    {\large Shubhaankar Gupta$^1$, Thomas P. O’Connell$^2$, Bernhard Egger$^{2,3}$}\\
    {\normalsize
    $^1$Delhi Public School RK Puram\\
    $^2$Massachusetts Institute of Technology\\
    $^3$Friedrich-Alexander-University Erlangen-Nuremberg\\
    guptashubhankar06@gmail.com
    }}
\begin{document}

\maketitle

\begin{abstract}
Pre-training on large-scale databases consisting of natural images and then fine-tuning them to fit the application at hand, or transfer-learning, is a popular strategy in computer vision. However, \textit{Kataoka et al., 2020} \cite{KataokaACCV2020} introduced a technique to eliminate the need for natural images in supervised deep learning by proposing a novel synthetic, formula-based method to generate 2D fractals as training corpus. Using one synthetically generated fractal for each class, they achieved transfer learning results comparable to models pre-trained on natural images. In this project, we take inspiration from their work and build on this idea-- using 3D procedural object renders. Since the image formation process in the natural world is based on its 3D structure, we expect pre-training with 3D mesh renders to provide an implicit bias leading to better generalization capabilities in a transfer learning setting and that invariances to 3D rotation and illumination are easier to be learned based on 3D data. Similar to the previous work, our training corpus will be fully synthetic and derived from simple procedural strategies; we will go beyond classic data augmentation and also vary illumination and pose which are controllable in our setting and study their effect on transfer learning capabilities in context to prior work. In addition, we will compare the 2D fractal and 3D procedural object networks to human and non-human primate brain data to learn more about the 2D vs. 3D nature of biological vision.
\end{abstract}

\section{Introduction}
Pre-training of artificial neural networks on available large-scale datasets offers a mechanism to reduce the need for excessive training data for the application at hand -- simply by transferring features from large-scale databases to datasets covering restricted domains with limited data. This allows models with small quantities of data to still fair exceptionally well in objective performance metrics. Pre-training has become increasingly popular in the computer vision community owing to a substantial rise in the availability of large-scale datasets like ImageNet \cite{ILSVRC15}. Since such natural image datasets have been widely accepted and lead to strong results with natural testing data, we might assume that images most closely resembling natural objects achieve maximum performance. However, since these natural images have to be manually labeled, their production is prone to errors and extremely labor-intensive. Recently, \cite{KataokaACCV2020} countered the notion that pre-training can only be performed on natural images and proposed pre-training on 2D fractals. For most datasets, such pre-training led to performance comparable to models pre-trained on ImageNet and for some it even exceeds their performance.

We propose a novel technique for pre-training with a strong 3D inductive bias, which we plan to pursue this by replacing 2D fractals with procedurally generated 3D meshes rendered under various illumination conditions and poses.
Rendering multiple images from a single mesh permits determining the type and magnitude of the variation to be implemented on a single object. It paves the way to generate massive numbers of images per class without manual labeling of images and the potential human error accompanying it. The dataset produced will then be used in a classification setting where the training objective is to identify the correct class for a fractal or 3D object independent of deformations or camera and illumination variation.
We expect the 3D nature of our generated images to lead to a different inductive bias than the 2D fractals and expect networks pre-trained on 3D objects to generalize better to complex visual tasks, including object detection.

Intermediate features in convolutional neural networks (CNNs) optimized for visual tasks bear a notable resemblance to the hierarchy of neural activity along the primate ventral visual stream, which supports object recognition  \cite{yamins2014performance}.  Since our network will be trained using 3D variations, a thought-provoking question arises: \textit{Do models trained on 3D renders better match neural activity along the primate ventral stream than models trained with 2D images}? We will pursue experiments to explicitly study this question, comparing the relative linear fits between 2D/3D model features and brain activity from monkeys and humans. In addition, our 3D data generation process enables generation of 2D data (by removing all pose variation) and even enables the removal of 3D cues (by removing all illumination variation), allowing us to disentangle 2D and 3D properties of the primate visual system. Giving precedent to the computer vision aspect of the research which is to achieve better transfer learning performance compared to the 2D approach, we will also study whether an inductive bias towards 3D better predicts brain activity.

Since the natural image formation process is 3D, we hypothesize that our method will learn better low- and high-level features which better generalize for visual learning tasks and also match the brain data better than the 2D fractal work of \cite{KataokaACCV2020}. We propose the following steps to pursue this study:

\begin{enumerate}
\item Generate synthetic object renders based on simple algorithms by attributing few meshes to each class and rendering them under various viewpoints and illumination conditions. We will use various combinations of datasets formed from 2D and 3D data.

\item Train and evaluate artificial neural networks using 2D and 3D datasets following the protocol of \cite{KataokaACCV2020} and focus on the transfer learning capabilities on a set of image classification tasks and measure the performance of our models.

\item Evaluate brain predictability of the networks pre-trained based on those datasets to investigate if a 3D inductive bias produces networks that better resembles brain activity.
\end{enumerate}

\section{Related work}
\label{gen_inst}

\subsection{Rendering and procedural meshes}
The generation of synthetic data is popular for various computer vision tasks e.g. \cite{gan2020threedworld,kortylewski2018training}. The goal of 3D rendering is usually to simulate the 3D world closely and accurately using high-quality assets (e.g. in \cite{wood2021fake}). The idea of domain randomization uses 3D assets, but starts to randomize texture and illumination \cite{tremblay2018training} leading to non-realistic appearance. In contrast, our approach builds on easy to generate assets based on procedural strategies. Today, fractals are the most popular procedural structures for pre-training -- most, however, are 2D. Procedural 3D meshes are commonly used in the entertainment industry e.g. to generate landscapes \cite{olsen2004realtime}.

\subsection{Pre-training datasets}
Conventionally, pre-training of CNNs has been performed on large-scale natural image datasets \cite{CIFAR-10, 5206848, zhou2017places}. As noted by \cite{huh2016makes}, this practice of first training a network to perform image classification on large-scale datasets (i.e. pre-training) and then transferring the extracted features for a completely new work (i.e. fine-tuning) has become the rule of thumb for solving a wide range of computer vision problems. 
The development of denser and deeper supervised CNN architectures \cite{AlexNet,he2016deep,densenet,Simonyan15} has unquestionably impacted pre-training \cite{AlexNet}. While such models function at the mere cost of more computational resources, they have substantially amplified performance metrics since they extract a huge number of perceptive features at various hierarchies and iterations.

Most approaches rely on large-scale real world datasets for pre-training. In contrast, \cite{KataokaACCV2020} proposed to solely train on  synthetically generated images which are derived from simple rules (fractals). Their method used one formula to generate one class, and used data augmentation to increase the number of images in the category. This approach performed very well in comparison to pre-training with natural-image datasets. Our method is different from it since we use 3D procedural and morphable models to generate different 3D objects as classes and render them under different viewpoints and illumination conditions to obtain multiple images per class.

\subsection{Computational models for biological vision}
CNN features explain appreciable variance in monkey and human neural activity, and the layer hierarchy in CNNs maps onto the ventral stream hierarchy, which underlies primate object recognition \cite{yamins2014performance}. These general findings have been replicated across species (monkeys, humans), imaging modalities (electrophysiology, fMRI, MEG, EEG), and stimulus types (objects, scenes, faces) \cite{yamins2016using, richards2019deep}.
Since we assume that the primate brain's neural tuning develops to capture the 3D structure of the world, we hypothesize that pre-training with the 3D dataset will produce CNN features that better explain activity in high level brain area than pre-training with 2D datasets like \cite{KataokaACCV2020}.

\section{Methodology}

\label{headings}

\begin{figure}[htp]
    \centering
    \includegraphics[width = 1.0\textwidth]{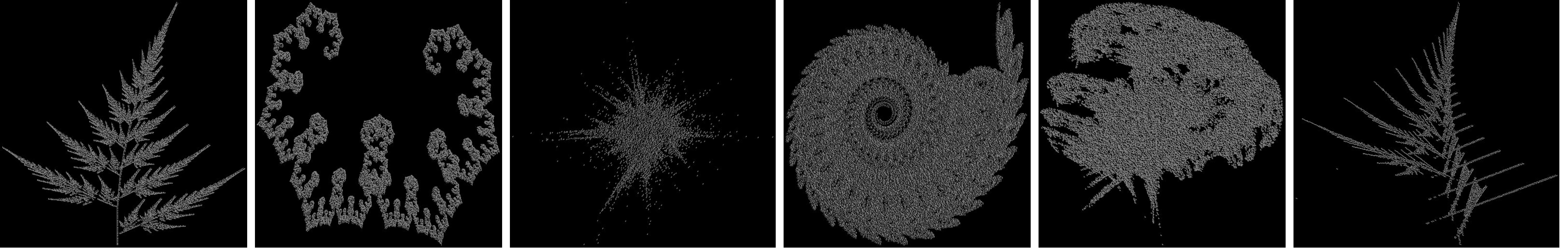}
    \caption{Fractal images from \textit{Kataoka et al., 2020} \cite{KataokaACCV2020}. This dataset is built using different fractal rules and image-based data augmentation. The dataset provides up to 10k different fractals (classes) and 1k images per class.}
    \label{fig:fractalDB}
\end{figure}

Our approach explores different combinations of synthetic datasets. We use FractalDB \cite{KataokaACCV2020} as a 2D baseline dataset (Fig.~\ref{fig:fractalDB}). Two 3D datasets will be built, namely ProcSynthDB (Fig.~\ref{fig:ProcSynthDB}) and MorphSynthDB (Fig.~\ref{fig:MorphSynthDB}). While the FractalDB uses a 2D fractal-based approach with 2D data augmentation for image generation, our proposed two synthetic 3D datasets build on three-dimensional procedural methods to generate 3D meshes per class and renders them under a variety of viewpoints and illumination conditions. A brief overview of the dataset comparisons can be found in Table \ref{table:overview}
We will add results of standard ImageNet trained networks (AlexNet, VGG, ResNet-50) as baseline to our comparison.

\subsection{ProcSynthDB}
\begin{figure}[htp]
    \centering
    \includegraphics[width = 1.0\textwidth]{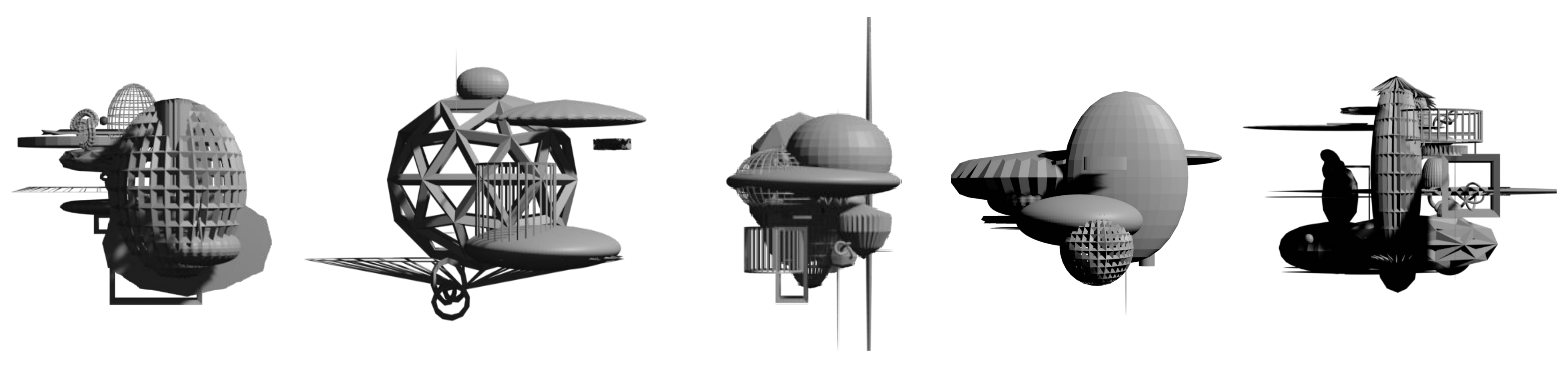}
    \caption{Example images of ProcSynthDB objects from different classes. In addition to the proceduraly created class, we change the viewpoint and illumination conditions.}
    \label{fig:ProcSynthDB}
\end{figure}

For every class, we start off by creating a base mesh from simple building blocks (sphere, cylinder, cube etc.) and using various wireframes, smootheness, skin on them. Each base mesh should be different enough to visually classified as belonging to a separate class. Our first approach uses perspective to create alterations in renders by changing the 3D viewpoint as well as the illumination conditions it is subjected to. This enables to get multiple images per mesh with a strong 3D-based variation in appearance.
In the rendering process, we also change the the size of the mesh twice across one single axis so that it produces natural-looking modifications. We describe the exact procedure to generate the meshes in pseudo-code in Algorithm~\ref{alg:two} and examples of resulting shapes are found in Fig.~\ref{fig:ProcSynthDB}.

\SetKwComment{Comment}{/* }{ */}
\RestyleAlgo{ruled}
\begin{algorithm}
\caption{Generation of ProcSynthDB Meshes}\label{alg:two}
\SetKwInOut{Input}{input}\SetKwInOut{Output}{output}
\Input{number of meshes $n$, $v=3$, $w=5$, $maxSize=10$}
\Output{$n$ synthetic 3D meshes}

$i \gets 0$ \;

\While{$i < n$}{
    \Comment{Add several primitives with random parameters and modifiers}
    $p \gets random(1,v)$ \;
        \For{$j\leftarrow 1$ \KwTo $p$}{
        \For{$objectType \in cube, sphere, cone, cylinder, torus$}{
            \For{$l\leftarrow 1$ \KwTo random(0,w)}{
                add $objecType$ with random parameters to the scene \;
                    randomly modify mesh with wireframe or subdivide modifier or not \;
            }
        }
    }
    \Comment{Quality Control to remove objects over the maximum size}
    \If{$max(size_x, size_y, size_z) 	\leq maxSize$}{
        $i \gets i+1$\
        save mesh from all objects \;
    }
    
    delete all objects \;

}
\end{algorithm}

\subsection{MorphSynthDB}
\begin{figure}[htp]
    \centering
    \includegraphics[width = 1.0\textwidth]{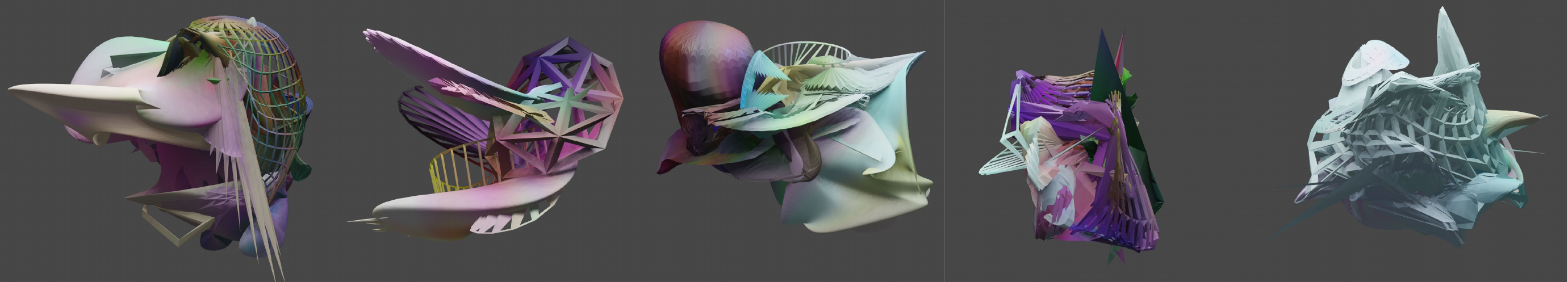}
    \caption{Example images of MorphSynthDB objects from different classes. In addition to shape variation based on Gaussian processes, this dataset also includes texture variation.}
    \label{fig:MorphSynthDB}
\end{figure}

Our second 3D render dataset builds on ProcSynthDB and explores Gaussian process deformation models \cite{luthi2017gaussian} for shape and texture variation \cite{sutherland2020building}. We will start from the first 100 ProcSynthDB meshes and add random shape and color variation to them to obtain new shapes with additional variation in color. We use the implementation of \cite{sutherland2020building} but with higher strength of the shape variation for the shape Gaussian processes. In particular we changed the magnitude of the following shape and albedo parameters (original value in brackets) $b_s = 50 (5), c_s = 300 (3), b_a = 0.05 (0.01), c_a = 0.2 (0.01)$. Since this procedure is sensitive to large number of vertices we downsample the meshes before applying the transformations. Examples of resulting shapes are found in Fig.~\ref{fig:MorphSynthDB}.

\begin{table}[h!]
\renewcommand{\arraystretch}{1.4}
\begin{center}
\begin{tabular}{| c || c | c | c |} 
 \hline
 \textbf{Database} & \textbf{View Dimensionality} & \textbf{Colored} & \textbf{Illumination Variation}\\ [0.5ex] 
 \hline
 ProcSynthDB & 3D & $\times$ & $\checkmark$ \\ 
 \hline
 MorphSynthDB & 3D & $\checkmark$ & $\checkmark$ \\
 \hline
 FractalDB & 2D & $\times$ & $\times$ \\
 \hline
\end{tabular}
\vspace{0.5em}
\caption{Overview of database compositions}
\label{table:overview}
\end{center}
\end{table}


\subsection{Pre-training and subsequent fine-tuning}
We will use the ResNet-50 architecture \cite{he2016deep} for pre-training our models across all dataset combinations. Our process will stay restricted to supervised multi-class classification. We will change the training phase of the model with our dataset leaving the fine-tuning step untampered, as proposed in \cite{KataokaACCV2020}. For the sake of maintaining a uniform comparison, we will perform a hyperparameter search based on cross-validation for each set of network and data and will search around the values derived in \cite{KataokaACCV2020}.

\subsection{Comparing models to brain data}
To assess the similarity between features in the procedural 2D and 3D models and brain activity, we will use linear regression to predict neural responses from monkeys and humans from model features \cite{kay2008identifying}. The general procedure is as follows: 1.) model features are computed for the images used to measure neural responses, 2.) using 10-fold cross-validation, partial least squares regression will be used to learn a linear mapping from model features to neural activity, and 3.) using the held-out data, the overall fit between a given model and neural responses will be quantified as the correlation between the predicted and true neural responses \cite{yamins2014performance}. This final correlation between true and predicted responses constitutes the \textit{neural predictivity} of a given model for the targeted brain region. We will also perform Representational Similarity Analysis (RSA) to measure the similarity between our computational model and brain activity.

\section{Experimental protocol}

\subsection{Combinations of the pre-training datasets}

For our experiments, we pre-train our models using the two 3D render databases we produced i.e. the ProcSynthDB and MorphSynthDB; and the third 2D baseline FractalDB. We will carry out our experiments on varied combinations achieved by incorporating these three datasets by utilizing six permutations of our them, as shown in Table~\ref{table:datasets}.

\begin{table}[h!]
\small
\renewcommand{\arraystretch}{1.4}
\begin{center}
\begin{tabular}{|l|p{2.5cm}|} 
 \hline
 \textbf{Dataset combinations for pre-training} & \textbf{No. classes from each subdataset}\\ [0.5ex] 
 \hline
 FractalDB & 1000\\ 
 ProcSynthDB & 1000\\
 MorphSynthDB & 1000\\
 ProcSynthDB +  FractalDB & 500 \\
 ProcSynthDB + MorphSynthDB & 500\\
 ProcSynthDB + MorphSynthDB + FractalDB & 333\\ 
 \hline
\end{tabular}
\vspace{0.5em}
\caption{Combinations of the three datasets to be used during pre-training. We aim at a total of 1000 classes for each combination and will mix them uniformly. For each class we will generate 1000 images following the protocol of FractalDB \cite{KataokaACCV2020}.}
\label{table:datasets}
\end{center}
\end{table}

\cite{KataokaACCV2020} re-iterated the well-established fact that the performance achieved by pre-training models is in proportion to an increase in the number of classes and the number of objects or images per class. Therefore, for the sake of uniform recordings and to maintain a category-instance ration, we use 1000 category × 1000 instances per category for all datasets and combinations.

\subsection{Transfer learning}
We will pre-train ResNet-50 \cite{he2016deep} using all dataset combinations defined in Table~\ref{table:datasets}. Then, we will fine-tune the models on secondary tasks (natural image datasets, Table~\ref{table:tasks}) to assess their transfer learning performance. For the transfer learning experiments we will closely follow the protocol of \cite{KataokaACCV2020}. Better transfer performance after pre-training on any of our 3D object datasets relative to the 2D fractal datasets would support our hypothesis that pre-training with a strong 3D inductive bias will more closely capture the statistics of natural images.
In addition to this quantitative evaluation we will also visualize the learned features of the resulting models similar as in \cite{KataokaACCV2020} and compare them to features derived from ImageNet trained models.

\begin{table}[!h]
\small
\centering
\renewcommand{\arraystretch}{1.4}

\begin{tabular}{|c|c|c|c|c|c|c|} 
\hline
 \textbf{Test datasets}&
 CIFAR 10&
 CIFAR 100&
 ImageNet 1K&
 Places 365&
 Pascal VOC 2012&
 Omniglot\\
\hline
\end{tabular}
\vspace{0.5em}
\caption{We will use these datasets for our transfer learning evaluation of the models trained with the datasets in Table~\ref{table:datasets}. Those testing datasets are also used in the FractalDB work \cite{KataokaACCV2020}.}

\label{table:tasks}
\end{table}

\subsection{Comparing models to brain data}
The monkey neural responses \cite{Majaj13402} were measured for a set of 2560 images of rendered objects with multi-array electrophysiology from areas V4 (an intermediate visual area) and IT (a late-stage ventral stream area that represents high-level object properties). For this data, we predict that 2D models will better predict V4 responses, while 3D models will better predict IT responses.

The human neural responses \cite{Allen2021.02.22.432340} were measured for a large set of 10k+ natural images per subject with functional Magnetic Resonance Imaging (fMRI). For these analyses, we will target early (V1, V2), mid (V3, V4), and late (LO, PfS) stages of the ventral stream. We predict that the 2D models will
best predict responses in early- and mid-level regions, while the 3D models will best predict responses in late regions.

\subsection{Ablation study}
One of the key concerns of our method could be that the observed differences could be caused by the different techniques to generate the datasets. To overcome this limitation we will generate a 3D based dataset that has all the 3D variation removed - it will not contain illumination and pose variation and will therefore be a 2D dataset based on 3D data. For this ablation study we will ablate the best performing 3D dataset and build such a 2D variant and name it FlatWorldDB.

\section{Conclusion}

We propose a strategy to perform supervised classification using synthetic images rendered from 3D meshes. We will investigate the influence of 3D vs. 2D inductive bias via varying the set of fully synthetic databases. Additionally, we will compare the effects of the data-generation methods on a CNN's performance after transfer learning, on brain predictivity, and their similarity to each other. The goal of this research is not to outperform ImageNet trained models but to measure relative differences resulting from 2D vs. 3D pre-training.

\bibliographystyle{unsrtnat}
\bibliography{main}

\begin{thebibliography}{23}
\providecommand{\natexlab}[1]{#1}
\providecommand{\url}[1]{\texttt{#1}}
\expandafter\ifx\csname urlstyle\endcsname\relax
  \providecommand{\doi}[1]{doi: #1}\else
  \providecommand{\doi}{doi: \begingroup \urlstyle{rm}\Url}\fi

\bibitem[Kataoka et~al.(2020)Kataoka, Okayasu, Matsumoto, Yamagata, Yamada,
  Inoue, Nakamura, and Satoh]{KataokaACCV2020}
Hirokatsu Kataoka, Kazushige Okayasu, Asato Matsumoto, Eisuke Yamagata, Ryosuke
  Yamada, Nakamasa Inoue, Akio Nakamura, and Yutaka Satoh.
\newblock Pre-training without natural images.
\newblock \emph{Asian Conference on Computer Vision (ACCV)}, 2020.

\bibitem[Russakovsky et~al.(2015)Russakovsky, Deng, Su, Krause, Satheesh, Ma,
  Huang, Karpathy, Khosla, Bernstein, Berg, and Fei-Fei]{ILSVRC15}
Olga Russakovsky, Jia Deng, Hao Su, Jonathan Krause, Sanjeev Satheesh, Sean Ma,
  Zhiheng Huang, Andrej Karpathy, Aditya Khosla, Michael Bernstein,
  Alexander~C. Berg, and Li~Fei-Fei.
\newblock {ImageNet Large Scale Visual Recognition Challenge}.
\newblock \emph{International Journal of Computer Vision (IJCV)}, 115\penalty0
  (3):\penalty0 211--252, 2015.
\newblock \doi{10.1007/s11263-015-0816-y}.

\bibitem[Yamins et~al.(2014)Yamins, Hong, Cadieu, Solomon, Seibert, and
  DiCarlo]{yamins2014performance}
Daniel~LK Yamins, Ha~Hong, Charles~F Cadieu, Ethan~A Solomon, Darren Seibert,
  and James~J DiCarlo.
\newblock Performance-optimized hierarchical models predict neural responses in
  higher visual cortex.
\newblock \emph{Proceedings of the national academy of sciences}, 111\penalty0
  (23):\penalty0 8619--8624, 2014.

\bibitem[Gan et~al.(2020)Gan, Schwartz, Alter, Schrimpf, Traer, De~Freitas,
  Kubilius, Bhandwaldar, Haber, Sano, et~al.]{gan2020threedworld}
Chuang Gan, Jeremy Schwartz, Seth Alter, Martin Schrimpf, James Traer, Julian
  De~Freitas, Jonas Kubilius, Abhishek Bhandwaldar, Nick Haber, Megumi Sano,
  et~al.
\newblock Threedworld: A platform for interactive multi-modal physical
  simulation.
\newblock \emph{arXiv preprint arXiv:2007.04954}, 2020.

\bibitem[Kortylewski et~al.(2018)Kortylewski, Schneider, Gerig, Egger,
  Morel-Forster, and Vetter]{kortylewski2018training}
Adam Kortylewski, Andreas Schneider, Thomas Gerig, Bernhard Egger, Andreas
  Morel-Forster, and Thomas Vetter.
\newblock Training deep face recognition systems with synthetic data.
\newblock \emph{arXiv preprint arXiv:1802.05891}, 2018.

\bibitem[Wood et~al.(2021)Wood, Baltrusaitis, Hewitt, Dziadzio, Cashman, and
  Shotton]{wood2021fake}
Erroll Wood, Tadas Baltrusaitis, Charlie Hewitt, Sebastian Dziadzio, Thomas~J
  Cashman, and Jamie Shotton.
\newblock Fake it till you make it: Face analysis in the wild using synthetic
  data alone.
\newblock In \emph{Proceedings of the IEEE/CVF International Conference on
  Computer Vision}, pages 3681--3691, 2021.

\bibitem[Tremblay et~al.(2018)Tremblay, Prakash, Acuna, Brophy, Jampani, Anil,
  To, Cameracci, Boochoon, and Birchfield]{tremblay2018training}
Jonathan Tremblay, Aayush Prakash, David Acuna, Mark Brophy, Varun Jampani, Cem
  Anil, Thang To, Eric Cameracci, Shaad Boochoon, and Stan Birchfield.
\newblock Training deep networks with synthetic data: Bridging the reality gap
  by domain randomization.
\newblock In \emph{Proceedings of the IEEE conference on computer vision and
  pattern recognition workshops}, pages 969--977, 2018.

\bibitem[Olsen(2004)]{olsen2004realtime}
Jacob Olsen.
\newblock Realtime procedural terrain generation.
\newblock 2004.

\bibitem[Krizhevsky et~al.()Krizhevsky, Nair, and Hinton]{CIFAR-10}
Alex Krizhevsky, Vinod Nair, and Geoffrey Hinton.
\newblock Cifar-10 (canadian institute for advanced research).
\newblock URL \url{http://www.cs.toronto.edu/~kriz/cifar.html}.

\bibitem[Deng et~al.(2009)Deng, Dong, Socher, Li, Li, and Fei-Fei]{5206848}
Jia Deng, Wei Dong, Richard Socher, Li-Jia Li, Kai Li, and Li~Fei-Fei.
\newblock Imagenet: A large-scale hierarchical image database.
\newblock In \emph{2009 IEEE Conference on Computer Vision and Pattern
  Recognition}, pages 248--255, 2009.
\newblock \doi{10.1109/CVPR.2009.5206848}.

\bibitem[Zhou et~al.(2017)Zhou, Lapedriza, Khosla, Oliva, and
  Torralba]{zhou2017places}
Bolei Zhou, Agata Lapedriza, Aditya Khosla, Aude Oliva, and Antonio Torralba.
\newblock Places: A 10 million image database for scene recognition.
\newblock \emph{IEEE Transactions on Pattern Analysis and Machine
  Intelligence}, 2017.

\bibitem[Huh et~al.(2016)Huh, Agrawal, and Efros]{huh2016makes}
Minyoung Huh, Pulkit Agrawal, and Alexei~A Efros.
\newblock What makes imagenet good for transfer learning?
\newblock \emph{arXiv preprint arXiv:1608.08614}, 2016.

\bibitem[Krizhevsky et~al.(2012)Krizhevsky, Sutskever, and Hinton]{AlexNet}
Alex Krizhevsky, Ilya Sutskever, and Geoffrey~E. Hinton.
\newblock Imagenet classification with deep convolutional neural networks.
\newblock In \emph{Proceedings of the 25th International Conference on Neural
  Information Processing Systems - Volume 1}, NIPS'12, page 1097–1105, Red
  Hook, NY, USA, 2012. Curran Associates Inc.

\bibitem[He et~al.(2016)He, Zhang, Ren, and Sun]{he2016deep}
Kaiming He, Xiangyu Zhang, Shaoqing Ren, and Jian Sun.
\newblock Deep residual learning for image recognition.
\newblock In \emph{Proceedings of the IEEE conference on computer vision and
  pattern recognition}, pages 770--778, 2016.

\bibitem[Huang et~al.(2017)Huang, Liu, Van Der~Maaten, and
  Weinberger]{densenet}
Gao Huang, Zhuang Liu, Laurens Van Der~Maaten, and Kilian~Q Weinberger.
\newblock Densely connected convolutional networks.
\newblock In \emph{Proceedings of the IEEE conference on computer vision and
  pattern recognition}, pages 4700--4708, 2017.

\bibitem[Simonyan and Zisserman(2015)]{Simonyan15}
Karen Simonyan and Andrew Zisserman.
\newblock Very deep convolutional networks for large-scale image recognition.
\newblock In \emph{International Conference on Learning Representations}, 2015.

\bibitem[Yamins and DiCarlo(2016)]{yamins2016using}
Daniel~LK Yamins and James~J DiCarlo.
\newblock Using goal-driven deep learning models to understand sensory cortex.
\newblock \emph{Nature neuroscience}, 19\penalty0 (3):\penalty0 356--365, 2016.

\bibitem[Richards et~al.(2019)Richards, Lillicrap, Beaudoin, Bengio, Bogacz,
  Christensen, Clopath, Costa, de~Berker, Ganguli, et~al.]{richards2019deep}
Blake~A Richards, Timothy~P Lillicrap, Philippe Beaudoin, Yoshua Bengio, Rafal
  Bogacz, Amelia Christensen, Claudia Clopath, Rui~Ponte Costa, Archy
  de~Berker, Surya Ganguli, et~al.
\newblock A deep learning framework for neuroscience.
\newblock \emph{Nature neuroscience}, 22\penalty0 (11):\penalty0 1761--1770,
  2019.

\bibitem[L{\"u}thi et~al.(2017)L{\"u}thi, Gerig, Jud, and
  Vetter]{luthi2017gaussian}
Marcel L{\"u}thi, Thomas Gerig, Christoph Jud, and Thomas Vetter.
\newblock Gaussian process morphable models.
\newblock \emph{IEEE transactions on pattern analysis and machine
  intelligence}, 40\penalty0 (8):\penalty0 1860--1873, 2017.

\bibitem[Sutherland et~al.(2021)Sutherland, Egger, and
  Tenenbaum]{sutherland2020building}
Skylar Sutherland, Bernhard Egger, and Joshua Tenenbaum.
\newblock Building 3d morphable models from a single scan.
\newblock In \emph{Proceedings of the IEEE/CVF International Conference on
  Computer Vision}, pages 2514--2524, 2021.

\bibitem[Kay et~al.(2008)Kay, Naselaris, Prenger, and
  Gallant]{kay2008identifying}
Kendrick~N Kay, Thomas Naselaris, Ryan~J Prenger, and Jack~L Gallant.
\newblock Identifying natural images from human brain activity.
\newblock \emph{Nature}, 452\penalty0 (7185):\penalty0 352--355, 2008.

\bibitem[Majaj et~al.(2015)Majaj, Hong, Solomon, and DiCarlo]{Majaj13402}
Najib~J. Majaj, Ha~Hong, Ethan~A. Solomon, and James~J. DiCarlo.
\newblock Simple learned weighted sums of inferior temporal neuronal firing
  rates accurately predict human core object recognition performance.
\newblock \emph{Journal of Neuroscience}, 35\penalty0 (39):\penalty0
  13402--13418, 2015.
\newblock ISSN 0270-6474.
\newblock \doi{10.1523/JNEUROSCI.5181-14.2015}.
\newblock URL \url{https://www.jneurosci.org/content/35/39/13402}.

\bibitem[Allen et~al.(2021)Allen, St-Yves, Wu, Breedlove, Dowdle, Caron,
  Pestilli, Charest, Hutchinson, Naselaris, and Kay]{Allen2021.02.22.432340}
Emily~J. Allen, Ghislain St-Yves, Yihan Wu, Jesse~L. Breedlove, Logan~T.
  Dowdle, Brad Caron, Franco Pestilli, Ian Charest, J.~Benjamin Hutchinson,
  Thomas Naselaris, and Kendrick Kay.
\newblock A massive 7t fmri dataset to bridge cognitive and computational
  neuroscience.
\newblock \emph{bioRxiv}, 2021.
\newblock \doi{10.1101/2021.02.22.432340}.
\newblock URL
  \url{https://www.biorxiv.org/content/early/2021/02/22/2021.02.22.432340}.

\end{thebibliography}

\end{document}